\documentclass[10pt,twocolumn,letterpaper]{article}

\usepackage{cvpr}
\usepackage{times}
\usepackage{epsfig}
\usepackage{bm}
\usepackage[normalem]{ulem}
\usepackage{todonotes}

\usepackage{floatrow}
\newfloatcommand{capbtabbox}{table}[][\FBwidth]

\usepackage{booktabs}       
\usepackage{nicefrac}       
\usepackage{multirow}
\usepackage{amsmath}
\usepackage{amssymb}
\usepackage{array}

\newcommand{\bbox}{b^{(i)}}
\newcommand{\hatbox}{\hat{b}^{(i)}}

\newcommand{\mean}{\mathit{mean}}
\newcommand{\std}{\mathit{std}}

\cvprfinalcopy 

\def\cvprPaperID{5} 

\ifcvprfinal\fi

\title{MetaDetect: Uncertainty Quantification and Prediction Quality Estimates for Object Detection}

\author{Marius Schubert$^{1}$, Karsten Kahl$^{1}$ and Matthias Rottmann$^{1}$ \\
\\
$^{1}$University of Wuppertal, School of Mathematics and Natural Sciences\\
\tt\small{
\{%
\href{mailto:schubert@math.uni-wuppertal.de}{schubert},%
\href{mailto:kkahl@math.uni-wuppertal.de}{kkahl},%
\href{mailto:rottmann@math.uni-wuppertal.de}{rottmann}%
\}@math.uni-wuppertal.de}%
}

\usepackage{graphicx}
\usepackage{xcolor}
\usepackage{caption,subcaption}

\usepackage{hyperref}
\hypersetup{pagebackref=true,breaklinks=true,colorlinks,bookmarks=false}

\usepackage[nameinlink]{cleveref}
\DeclareMathOperator*{\argmax}{arg\, max}

\newcommand{\IoU}{\mathit{IoU}}

\usepackage{enumerate}

\begin{document}
\maketitle

\begin{abstract}
In object detection with deep neural networks, the box-wise objectness score tends to be overconfident, sometimes even indicating high confidence in presence of inaccurate predictions. 
Hence, the reliability of the prediction and therefore reliable uncertainties are of highest interest.
In this work, we present a post processing method that for any given neural network provides  predictive uncertainty estimates and quality estimates. 
These estimates are learned by a post processing model that receives as input a hand-crafted set of transparent metrics in form of a structured dataset. Therefrom, we learn two tasks for predicted bounding boxes. We discriminate between true positives ($\IoU\geq0.5$) and false positives ($\IoU < 0.5$) which we term meta classification, and we predict $\IoU$ values directly which we term meta regression. The probabilities of the meta classification model aim at learning the probabilities of success and failure and therefore provide a modelled predictive uncertainty estimate. On the other hand, meta regression gives rise to a quality estimate.
In numerical experiments, we use the publicly available YOLOv3 network and the Faster-RCNN network and evaluate meta classification and regression performance on the Kitti, Pascal VOC and COCO datasets.
We demonstrate that our metrics are indeed well correlated with the $\IoU$.
For meta classification we obtain classification accuracies of up to $98.92\%$ and AUROCs of up to $99.93\%$. For meta regression we obtain an $R^2$ value of up to $91.78\%$. These results yield significant improvements compared to other network's objectness score and other baseline approaches. Therefore, we obtain more reliable uncertainty and quality estimates which is particularly interesting in the absence of ground truth.

\vfill

\end{abstract}

\begin{figure*}[t]
\begin{minipage}[c]{.99\textwidth}
\begin{center}
\includegraphics[width=.13\textwidth]{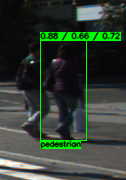}
\includegraphics[width=.13\textwidth]{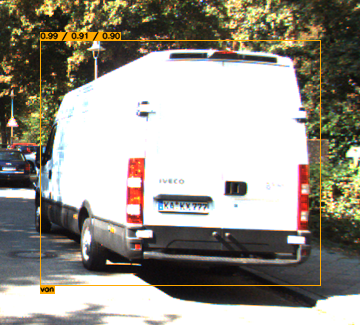}
\includegraphics[width=.13\textwidth]{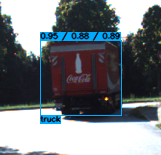}
\includegraphics[width=.13\textwidth]{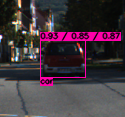}
\includegraphics[width=.13\textwidth]{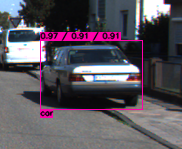}
\includegraphics[width=.13\textwidth]{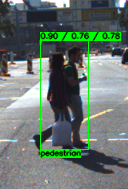}
\includegraphics[width=.13\textwidth]{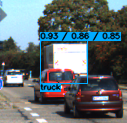} \\
\includegraphics[width=.13\textwidth]{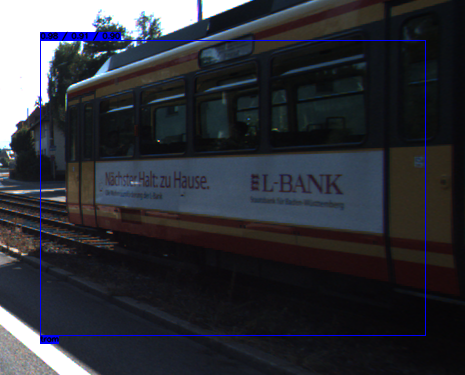}
\includegraphics[width=.13\textwidth]{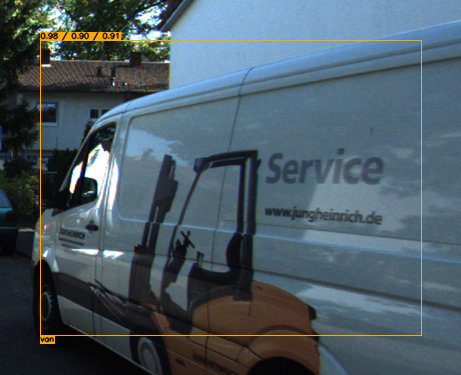}
\includegraphics[width=.13\textwidth]{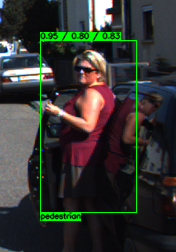}
\includegraphics[width=.13\textwidth]{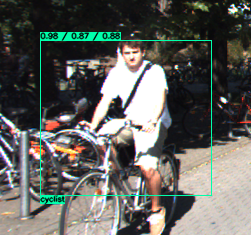}
\includegraphics[width=.13\textwidth]{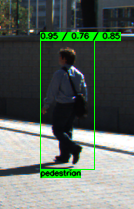}
\includegraphics[width=.13\textwidth]{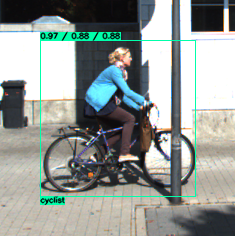}
\includegraphics[width=.13\textwidth]{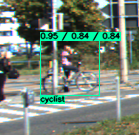}
\end{center}
\captionof{figure}{Examples of predicted bounding boxes with objectness score / true $\IoU$ / predicted $\IoU$. The predictions, and therefore the data to train and evaluate gradient boosting, are made with the YOLOv3 network, the Kitti dataset and a score threshold $t=0.01$. \label{fig:Collage_Regression}}
\end{minipage}
\end{figure*}

\section{Introduction} \label{sec:intro}
In recent years, deep neural networks have surpassed other approaches in many tasks with unstructured data. However their in-transparent nature poses many questions and problems.
In particular in safety-critical applications, the reliability of deep neural network predictions is of highest interest as neural networks tend to provide high confidences even when they fail \cite{Goodfellow14Adversarial}. In order to detect these and therefore make neural networks more interpretable, meaningful uncertainties are required \cite{Bayesian_Framework}.
%
A broad survey on uncertainty in machine learning can be found in~\cite{Overview}.
A very important and also reducible type of uncertainty is the model uncertainty resulting from the fact that the ideal parameters are unknown and have to be estimated from data \cite{Bayesian_Framework,Varitational_Bayesian,MC-Dropout,Early_Stopping}.
Bayesian models consider these uncertainties~\cite{Bayesian_Framework}. However, as Bayesian models for deep learning are nowadays infeasible, many different frameworks based on variational approximations have been proposed~\cite{Varitational_Bayesian,Early_Stopping,MC-Dropout}. 
For instance, Monte-Carlo (MC) Dropout~\cite{MC-Dropout} is used as a feasible approximation to Bayesian inference.

In classification tasks, a natural approach to detect incorrect predictions / false positives is to
introduce a dispersion measure on the softmax probabilities, such as the entropy or one minus the highest softmax probability and then threshold on this \cite{Principled,A_Baseline}. As in the works \cite{rottmann_date20,MetaSeg,MetaFusion,NestedMetaSeg,Time_Dynamic_MetaSeg} that extend \cite{A_Baseline} to semantic segmentation, we refer to this task as \emph{meta classification}.
A good meta classification performance requires well adjusted predictive uncertainty measures. 
Variational approximations of Bayesian learning are one approach to this. However, also feasible ones like \cite{MC-Dropout} still require numerous inferences of one and the same input in order to estimate predictive uncertainty.
In object detection, the network's objectness score is subject to thresholding which in a natural sense can be considered built in meta classification as part of state-of-the-art object detection pipelines \cite{DBLP:journals/corr/abs-1804-02767,DBLP:journals/corr/RenHG015}.

Confidence calibration has been proposed to account for the miscalibration of the objectness score \cite{Calibrating,Guo2017}. However, most approaches (e.g.\ histogram-based ones \cite{Kueppers2020}) calibrate the score in a statistical sense. This is not sufficient to account for the correctness or quality of a single predicted bounding box. Therefore, uncertainty estimates for objects detection networks seem more promising.

In~\cite{Evaluating} a baseline for meta classification in object detection is presented. The localization variables are determined using the candidate boxes that are present after the score thresholding and before the non-maximum suppression (NMS). The mean values of the associated candidate boxes represent the predicted bounding box and the added variances represent the localization uncertainty. In combination with 
the class uncertainty, which is generated from the entropy of the class probabilities in a softmax output, an uncertainty measure is obtained for each box. 
A threshold is used to decide whether the box is accepted or rejected as a prediction. To generate the uncertainty not only from the output, in~\cite{Uncertainty_Estimation_One-Stage,Uncertainty_Estimation_Safety,Softer-NMS,BayesOD} the loss is changed in such a way that the uncertainties of the localization variables are learned and displayed in their own newly introduced neurons. 
In~\cite{Uncertainty_Estimation_One-Stage,BayesOD} MC-Dropout is used for the localization and variance calculation of the predicted bounding boxes. The uncertainty introduced in the loss combined with the localization variance is applied to the individual anchor boxes/priors. In~\cite{BayesOD} the NMS is replaced by a Bayesian inference, in particular the assignment is done with a different cluster method. The performance with respect to prediction accuracy for all the listed performance metrics (mAP, PDQ Score, mGMUE, mCMUE) increases 
compared to other state of the art methods introduced in~\cite{Uncertainty_Estimation_One-Stage,Uncertainty_Estimation_Safety,Dropout_Sampling,Evaluating,Towards_Safe_Autonomous_Driving,Leveraging_Heteroscedastic}.
In~\cite{Dropout_Sampling}, classification uncertainty is extracted
from object detection networks via dropout sampling.

For semantic segmentation, a number of works that estimate the segmentation quality have been introduced \cite{QualityNet,Brain_Segmentation,Propagating_Uncertainty,MetaSeg,NestedMetaSeg,Time_Dynamic_MetaSeg,MetaFusion}. These works make use of the richness of information of the segmentation networks output, as there is a probability distribution available in each pixel of a given segment. In object detection tasks, this richness is not present.


\paragraph{Our contribution.}
In this work we extend the false positive detection baseline of~\cite{A_Baseline} from classification to object detection.
As the works \cite{MetaSeg,NestedMetaSeg,Time_Dynamic_MetaSeg,MetaFusion} in semantic segmentation we consider two (meta) tasks:
Discrimination between true positive ($\IoU \geq 0.5$) and false positive ($\IoU < 0.5$) boxes which is termed \emph{meta classification} and regression with $\IoU$ values which is termed \emph{meta regression}.
To approach these tasks without significantly increasing the computational cost caused object detection inferences, we introduce a post processing framework that in general can be added to any object detection network. Only by means of the network's output, our framework trains two models, i.e., one for each meta task.
More precisely, we construct a set of handcrafted and interpretable metrics that may reveal the network's uncertainty about a prediction. Amongst others, this includes the number of candidate boxes before NMS that overlap with the given predicted box to a chosen extent, the score, the class probability distribution, the boxes size and aspect ratio and many others. If desired, these metrics can be extended by MC-dropout statistics. In general any object-wise metric that seems to be helpful can be added.
From this we obtain a structured dataset where the rows amount to the predicted objects in a given number of images and the columns amount to the constructed metrics. From this set of metrics we learn both meta tasks with different classical machine learning models, i.e., linear ones, shallow neural networks and gradient boosting.
\Cref{fig:Collage_Regression} illustrates the effect of meta regression. 
In the depicted cases, the bounding boxes are predicted with an overconfident score, whereas the true $\IoU$ is significantly lower and the predicted $\IoU$ is close to the true one.

In our numerical experiments, we study the correlation of our constructed metrics with the $\IoU$ of prediction and ground truth, we study different sets of metrics and compare those to a score baseline. For both meta tasks, we significantly outperform the baseline. This is observed consistently over different datasets (Kitti \cite{Geiger2012CVPR}, Pascal VOC \cite{Everingham15} and COCO \cite{COCO}) and different object detection networks (YOLOv3 \cite{DBLP:journals/corr/abs-1804-02767} and Faster-RCNN \cite{DBLP:journals/corr/RenHG015}). We do not observe an improvement when performing Monte-Carlo dropout with a modified YOLOv3 network and including the obtained uncertainties into our set of metrics. This strengthens the statement, that our approach is reliable.

\paragraph{Related work.}
In this section, we clarify the differences between related works and ours.
The idea behind this work is in spirit similar to the approaches for semantic segmentation \cite{QualityNet, Brain_Segmentation, Propagating_Uncertainty, MetaSeg,NestedMetaSeg,Time_Dynamic_MetaSeg, MetaFusion}, however the nature of the output provided by segmentation networks and object detection networks is so different, such that the resulting uncertainty metrics are also clearly different.

Our approach is solely based on post processing. One can plug-in MC dropout quantities if the network architecture allows, however, this is not mandatory. On the other hand, \cite{Uncertainty_Estimation_One-Stage, Uncertainty_Estimation_Safety, Softer-NMS, BayesOD} incorporate the uncertainty quantification into the original network and change their loss functions and also the ultimate layer. This requires additional training and does not aim at quantifying the uncertainty of the original network. In addition, as opposed to the other works we provide an simple and modular statistical benchmark suite that can be used for any object detection network in combination with any object-wise uncertainty quantification method to obtain performance indicators for the latter.


In~\cite{Evaluating} box-wise uncertainty information is calculated by means of associated candidate boxes.
Whether a prediction is accepted or rejected is decided by simply thresholding on the resulting measure.
Beyond that, neither meta regression, meta classification nor a general statistical evaluation of uncertainty measures is performed. Instead they use their uncertainty measure to improve the non maximum suppression. In~\cite{Calibrating} introduces a confidence calibration method that uses variances from a Bayesian neural network trained on the confidence calibration task. This aims at estimating the uncertainty, given the localization variables and a specific confidence level, whereas we estimate predictive uncertainty for a given input. Meta classification was not performed.
%
%
In~\cite{Dropout_Sampling} the authors extract classification, localization and spatial uncertainty under dropout sampling and turn this into an overall performance improvement. 
Also in that work, there is no meta classification or regression performed and the proposed concept requires dropout inference. 

\paragraph{Outline.}
The remainder of this work is structured as follows: In \cref{sec:object detection} we introduce the concept of object detection 
and explain how to perform uncertainty quantification by means of candidate boxes and MC dropout. This is followed by the construction of uncertainty metrics using predictive uncertainty and geometry information in \cref{sec:metrics}. Afterwards we present numerical results in \cref{sec:numexp}, benchmarking several approaches against each other. First, we present the chosen parameters and datasets we have used for our experiments with the YOLOv3 and Faster-RCNN networks. Afterwards we discuss correlation coefficients and evaluate the usefulness of our box-wise uncertainty metrics for meta classification and regression.

\section{The Generic Object Detection Pipeline} \label{sec:object detection}

In this section we briefly review the concept of object detection and its commonly used components, as they will be of interest for our uncertainty quantification method and experiments with those.

Any image $x$ from the training or validation data is equipped with a set $y$ containing $G$ ground truth bounding boxes, i.e.,
\begin{equation}
    y = \{ b^{(i)} : i=1,...,G \} \, ,
\end{equation}
where each ground truth box is a tuple
\begin{equation}
    b^{(i)} = (r^{(i)}_\mathrm{min},r^{(i)}_\mathrm{max},c^{(i)}_\mathrm{min},c^{(i)}_\mathrm{max},\kappa^{(i)})
\end{equation}
containing minimal and maximal row and column indices $r^{(i)}_\mathrm{min},r^{(i)}_\mathrm{max},c^{(i)}_\mathrm{min}$ and $c^{(i)}_\mathrm{max}$ as well as a class index $\kappa^{(i)} \in \{ 1,\ldots,C \}$.
Given an image $x$, a neural network $f$ computes a set of a fixed number $N$ of inferred bounding boxes (or candidate boxes)
\begin{equation}
    f(x) = \hat{y}_c = \{ \hat{b}^{(i)} : i=1,\ldots,N \}
\end{equation}
where each $\hat{b}^{(i)}$ is a tuple of estimated values
\begin{equation}
    \hat{b}^{(i)} = (\hat{r}^{(i)}_\mathrm{min},\hat{r}^{(i)}_\mathrm{max},\hat{c}^{(i)}_\mathrm{min},\hat{c}^{(i)}_\mathrm{max},s^{(i)},p^{(i)}_1,\ldots,p^{(i)}_C) \, .
\end{equation}
Therein, the initial four values have the same meaning as for the ground truth, $s^{(i)}$ denotes the so-called (objectness) score which takes values in $[0,1]$ and (roughly) indicates how likely it is that $b^{(i)}$ contains an object. Furthermore, given an input $x$ the network $f$ provides class probabilities $p^{(i)}_1,\ldots,p^{(i)}_C$. The predicted class is given by $\hat{\kappa}^{(i)} = \argmax_{k=1,\ldots,C} p^{(i)}_k$.

A typical performance measure, that indicates to which extent a prediction is in accordance to the ground truth, is the $\IoU$. By defining the sets of pixels contained in a ground truth box $b^{(i)}$, that is $\beta^{(i)} = \{ (r,c) : r^{(i)}_\mathrm{min} \leq r \leq r^{(i)}_\mathrm{max}, \; c^{(i)}_\mathrm{min} \leq c \leq c^{(i)}_\mathrm{max} \}$, and analogously defining $\hat{\beta}^{(j)}$ for a estimated box $\hat{b}^{(j)}$, the $\IoU$ of $\hat{b}^{(i)}$ and $b^{(j)}$ is defined by
\begin{equation}
    \IoU(\hat{b}^{(i)},b^{(j)}) = \frac{| \hat{\beta}^{(i)} \cap {\beta}^{(j)}  |}{|\hat{\beta}^{(i)} \cup {\beta}^{(j)} |} \, ,
\end{equation}
where $\hat{\kappa}^{(i)}=\kappa^{(j)}$.

Downstream of a typical object detection pipeline, the \emph{candidate boxes} $\hat{y}_c$ are filtered by discarding all boxes $\hat{b}^{(i)}$ whose corresponding estimated score values $s^{(i)}$ remain below a chosen threshold $t$. That is, we define \emph{filtered candidate boxes} 
\begin{equation}
    \hat{y}_s = \{ \hat{b}^{(i)} \in \hat{y}_c : \hat{s}^{(i)} \geq t \} \, .
\end{equation}
Typically this is followed by the so-called \emph{non-maximum suppression} (NMS). 

Let initially $\hat{y} = \emptyset$. Iteratively, the estimated box $\hat{b}^{(i)} \in \hat{y}_s$ with maximal score $s^{(i)}$ is determined and all other boxes $\hat{b}^{(j)}$ with $\hat{\kappa}^{(i)}=\hat{\kappa}^{(j)}$ and $\IoU(\hat{b}^{(i)},\hat{b}^{(j)}) \geq \tau$, where $\tau$ is a specific threshold, are discarded from $\hat{y}_s$. The box $\hat{b}^{(i)}$ is also removed from $\hat{y}_s$ and added to $\hat{y}$. This is repeated until $\hat{y}_s=\emptyset$.
Afterwards, $\hat{y}$ represents the set of predicted boxes. For a more precise description of specific object detection pipelines, we refer to~\cite{DBLP:journals/corr/RedmonDGF15,DBLP:journals/corr/abs-1804-02767,DBLP:journals/corr/RenHG015}.

\section{Uncertainty Metrics for Object Detection} \label{sec:metrics}


In this section we construct uncertainty metrics for every $\hatbox\in\hat{y}$. We do so in two stages, first by introducing the general metrics that can be obtained from the object detection pipeline as is. Second, we extend this by additional metrics that can be computed when using MC dropout.

We consider a predicted bounding box $\hatbox\in\hat{y}$ and its corresponding filtered candidate boxes $\hat{b}^{(j)} \in \hat{y}_s$, that were discarded by the NMS.
+
The number of corresponding candidate boxes $\hat{b}^{(j)} \in \hat{y}_s$ filtered by the NMS intuitively gives rise to the likelihood of observing a true positive. The more candidate boxes $\hat{b}^{(j)}$ belong to $\hat{b}^{(i)}$, the more likely it seems that $\hat{b}^{(i)}$ is a true positive. We denote by $N^{(i)}$ the number of candidate boxes $\hat{b}^{(j)}$ belonging to $\hat{b}^{(i)}$ but suppressed by NMS. 
We increment this number by $1$ and also count in $\hat{b}^{(i)}$.
%

For a given image $x$ we have the set of predicted bounding boxes $\hat{y}$ and the ground truth $y$. As we want to calculate values that represent the quality of the neural network's prediction, we first have to define uncertainty metrics for the predicted bounding boxes in $\hat{y}$. For each $\hatbox \in \hat{y}$, we define the following quantities:
\begin{itemize}
    \setlength\itemsep{-0.05em}
    \item the number of candidate boxes $N^{(i)} \geq 1$ that belong to $\hatbox$ ($\hatbox$ belongs to itself),
    \item the predicted box $\hatbox$, i.e., the values of the tuple $(\hat{r}^{(i)}_\mathrm{min},\hat{r}^{(i)}_\mathrm{max},\hat{c}^{(i)}_\mathrm{min},\hat{c}^{(i)}_\mathrm{max},s^{(i)},p^{(i)}_1,\ldots,p^{(i)}_C)\in\mathbb{R}^{5+C}$,
    \item the size $d=(\hat{r}^{(i)}_\mathrm{max}-\hat{r}^{(i)}_\mathrm{min})\cdot(\hat{c}^{(i)}_\mathrm{max}-\hat{c}^{(i)}_\mathrm{min})$ and the circumference $g=2\cdot(\hat{r}^{(i)}_\mathrm{max}-\hat{r}^{(i)}_\mathrm{min})+2\cdot(\hat{c}^{(i)}_\mathrm{max}-\hat{c}^{(i)}_\mathrm{min})$,
    \item $\IoU_{\mathit{pb}}$: the $\IoU$ of $\hatbox$ and the box with the second highest score that was suppressed by $\hatbox$. This values is zero if there are no boxes corresponding to $\hatbox$ suppressed by the NMS ($N^{(i)}=1$),
    \item the minimum, maximum, arithmetic mean and standard deviation for all $(\hat{r}^{(i)}_\mathrm{min},\hat{r}^{(i)}_\mathrm{max},\hat{c}^{(i)}_\mathrm{min},\hat{c}^{(i)}_\mathrm{max},s^{(i)})$, size $d$ and circumference $g$ from $\hatbox$ and all the filtered candidate boxes that were discarded from $\hatbox$ in the NMS,
    \item the minimum, maximum, arithmetic mean and standard deviation for the $\IoU$ of $\hatbox$ and all the candidate boxes corresponding to $\hatbox$ that suppressed in the NMS,
    \item relative sizes $rd=d/g$, $rd_{\min}=d/g_{\min}$, $rd_{\max}=d/g_{\max}$, $rd_{\mean}=d/g_{\mean}$ and $rd_{\std}=d/g_{\std}$,
    \item the $\IoU$ of $\hatbox$ and the ground truth $\bbox$, this is not an input to a meta model but serves as the ground truth provided to the respective loss function.
\end{itemize}
Altogether, this results in $46+C$ uncertainty metrics.

We now elaborate on how to calculate uncertainty metrics for every $\hatbox\in\hat{y}$ when using MC dropout.
To this end, we consider the bounding box $\hatbox\in\hat{y}$ that was predicted without dropout and then we observe under dropout $J$ times the output of the same anchor box that produced $\hatbox$ and denote them by $\hatbox_1,...,\hatbox_J$. For these $J+1$ boxes we calculate the minimum, the maximum, the arithmetic mean and the standard deviation for the localization variables and the objectness score over $\hatbox,\hatbox_1,...,\hatbox_J$. This is done for every $\hatbox\in\hat{y}$ and results in 20 additional dropout uncertainty metrics. This means $66+C$ uncertainty metrics in total. 
Executing this procedure for all available test images we end up with a structured dataset. The number of predicted objects constitutes to the number of rows and the columns are given by the registered metrics.
After defining a training / test splitting of this dataset, we learn meta classification ($\IoU\geq0.5$ vs. $\IoU <0.5$) and meta regression (quantitative $\IoU$ prediction) from the training part of this data. \\
All these presented metrics, except for the true $\IoU$ can be computed without the knowledge of the ground truth. Our aim is to analyze to which extent they are suited for the tasks of meta classification and meta regression.

\section{Numerical Experiments: Pascal VOC, KITTI and COCO} \label{sec:numexp}
In this section we investigate the properties of the metrics defined in the previous sections for the example of object detection for three different datasets and two different networks. We deploy the YOLOv3 network~\cite{DBLP:journals/corr/abs-1804-02767} for which we use a reference implementation in Tensorflow \cite{tensorflow2015-whitepaper} as well as weights self-trained on the Kitti dataset \cite{Geiger2012CVPR}, the Pascal VOC2007 dataset~\cite{Everingham15} and the COCO dataset~\cite{COCO}. For Kitti we split the labelled training images as for the test images are no labels available. The images 0-5480 are used to train our network and the last 2000 (image 5481-7480) are used to evaluate our method. For Pascal VOC we used the training images from the years 2007 to 2011 to train our network and we evaluate our method on the 4952 Pascal VOC2007 test images. For COCO we train on all COCO2014 training images and evaluate our method on 2500 randomly selected test images from the COCO2014 test images.
For the Faster-RCNN~\cite{DBLP:journals/corr/RenHG015} we use a reference implementation in Tensorflow as well as pre-trained weights for all three datasets. When evaluating our method on the Kitti dataset we use all available labelled images. The Faster-RCNN is trained exclusively for the two classes ``person'' and ``car''. Eventually, the Faster-RCNN might have overfitted the training data. Indeed we observe high prediction accuracy. For Pascal VOC and COCO we use the same images as for our tests with the YOLOv3 network.

For more information about the default training and test parameters that we do not deviate from, we refer to the publicly available source codes and to \cite{DBLP:journals/corr/RedmonDGF15, DBLP:journals/corr/abs-1804-02767, DBLP:journals/corr/RenHG015} for a detailed explanation of the respective parameters.

We evaluate our methods for meta classification and regression on 33 different score thresholds, starting at 0.01, continuing with thresholds equal to $k/40$ until reaching 0.8 ($k=1,\ldots,32$). Over the course of thresholds, the highest mean average precision (mAP) values obtained by YOLOv3 are $91.99\%$ on Kitti, $86.9\%$ on Pascal VOC and $61.13\%$ on COCO. For a given class, the average precision (AP) is the area under precision recall curve. Mean average precision is the mean of the AP values for all considered classes. For all three datasets the mAP is equal to the Pascal VOC metric~\cite{mAP}, which determines true and false positives by means of an $\IoU$ threshold equal to $0.5$ (AP@.5), which is consistent with our definition of meta classification. 
For the Faster-RCNN we evaluate our method also on these 33 different score thresholds, with one  exception which is the Kitti dataset. The classes ``person'' and ``car'' are often predicted with extremly confident scores (very close to 1 or 0) such that the predictions differ only marginally at thresholds near 0.8 and 0.1. Therefore, thresholds of $10^{-1},\ldots,10^{-12}$ were chosen were chosen for the KITTI dataset. This leads to highest obtained mAP values for Faster-RCNN which are $89.33\%$ on Kitti, $80.31\%$ on Pascal VOC and $55.29\%$ on COCO.

\begin{figure*}[t]
    \centering 
\begin{subfigure}{0.33\textwidth}
  \includegraphics[width=\textwidth]{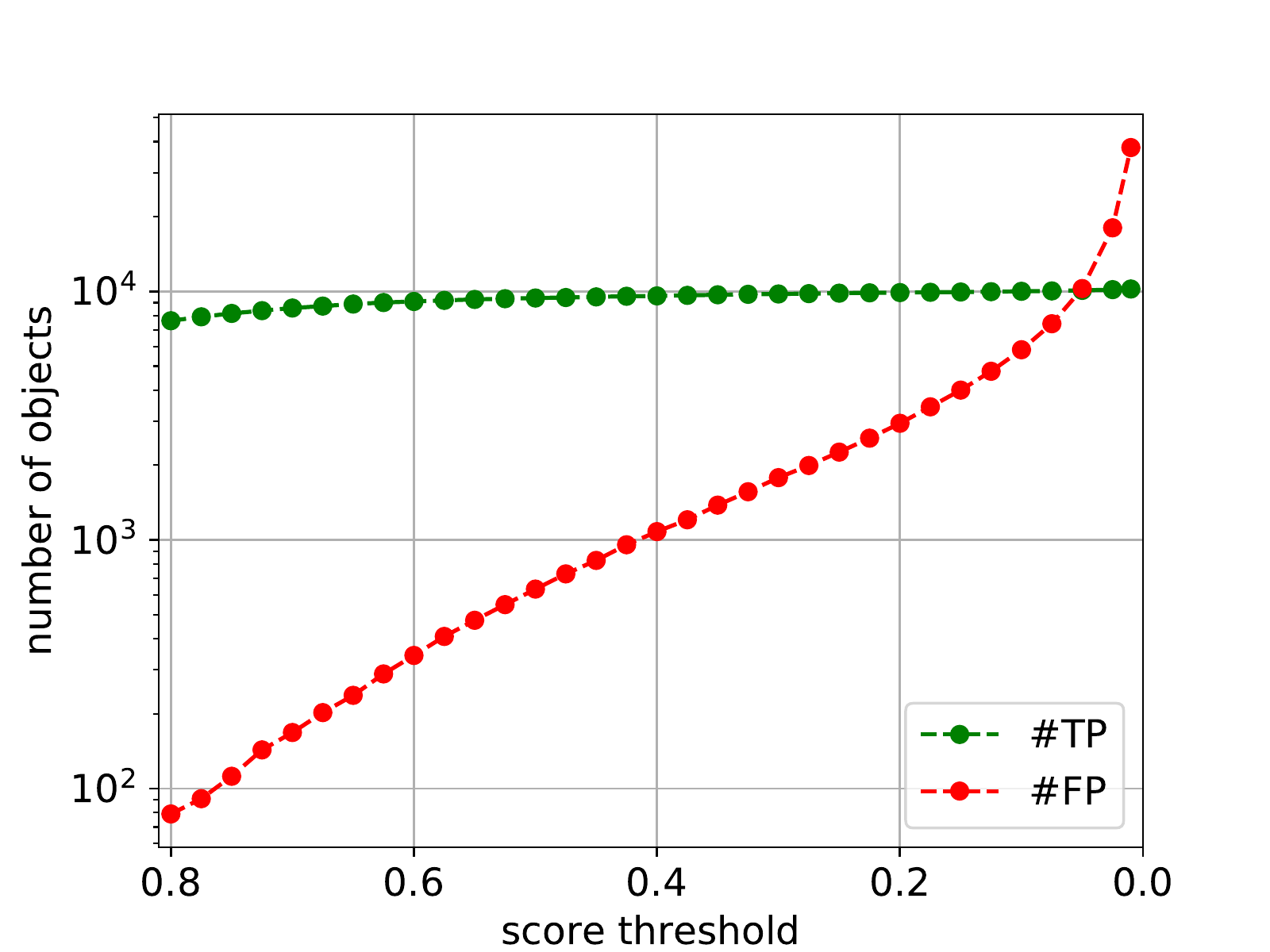}
  \caption{Kitti + YOLOv3}
  \label{fig:1}
\end{subfigure}\hfil 
\begin{subfigure}{0.33\textwidth}
  \includegraphics[width=\textwidth]{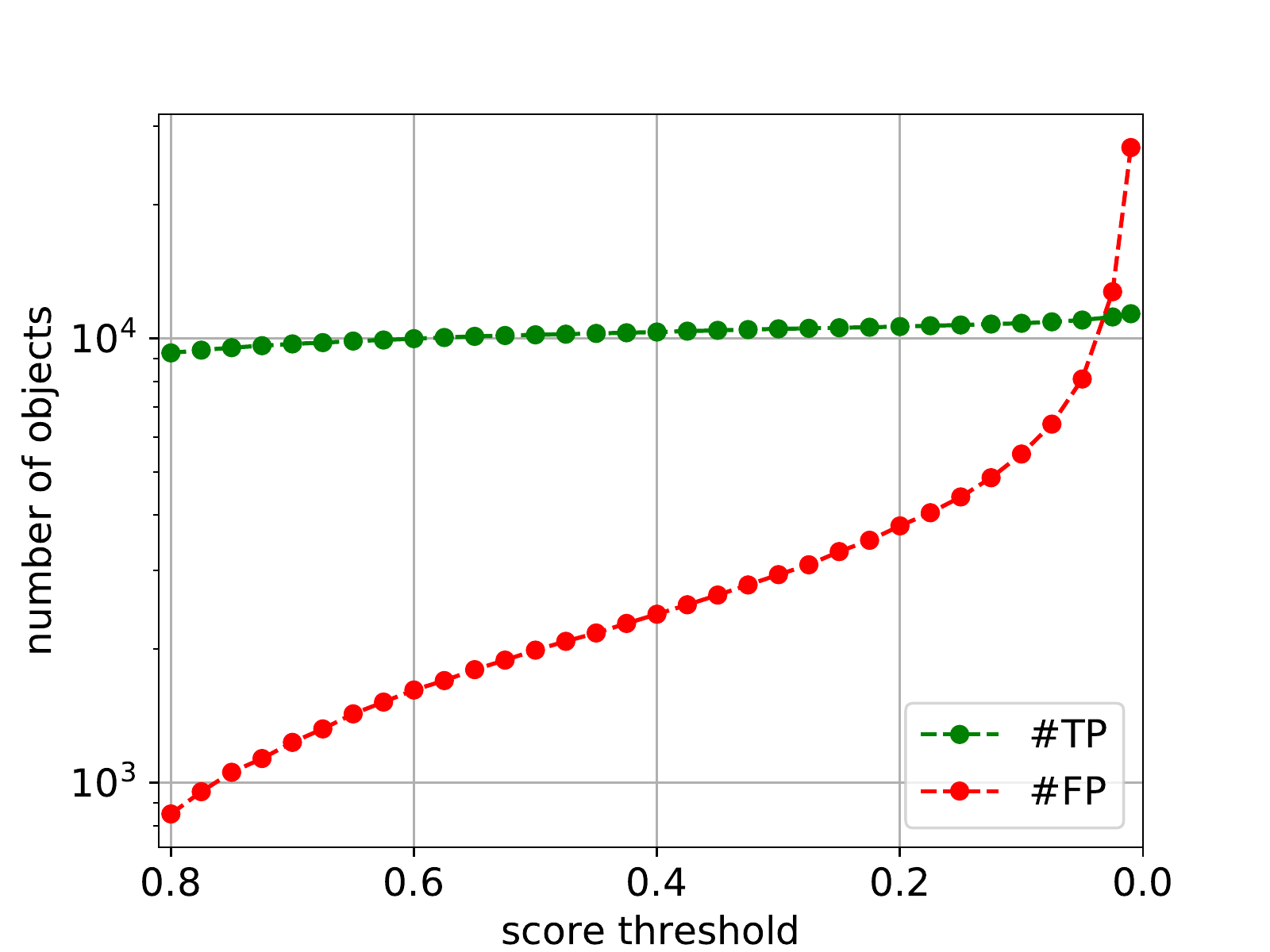}
  \caption{Pascal VOC + YOLOv3}
  \label{fig:2}
\end{subfigure}\hfil 
\begin{subfigure}{0.33\textwidth}
  \includegraphics[width=\textwidth]{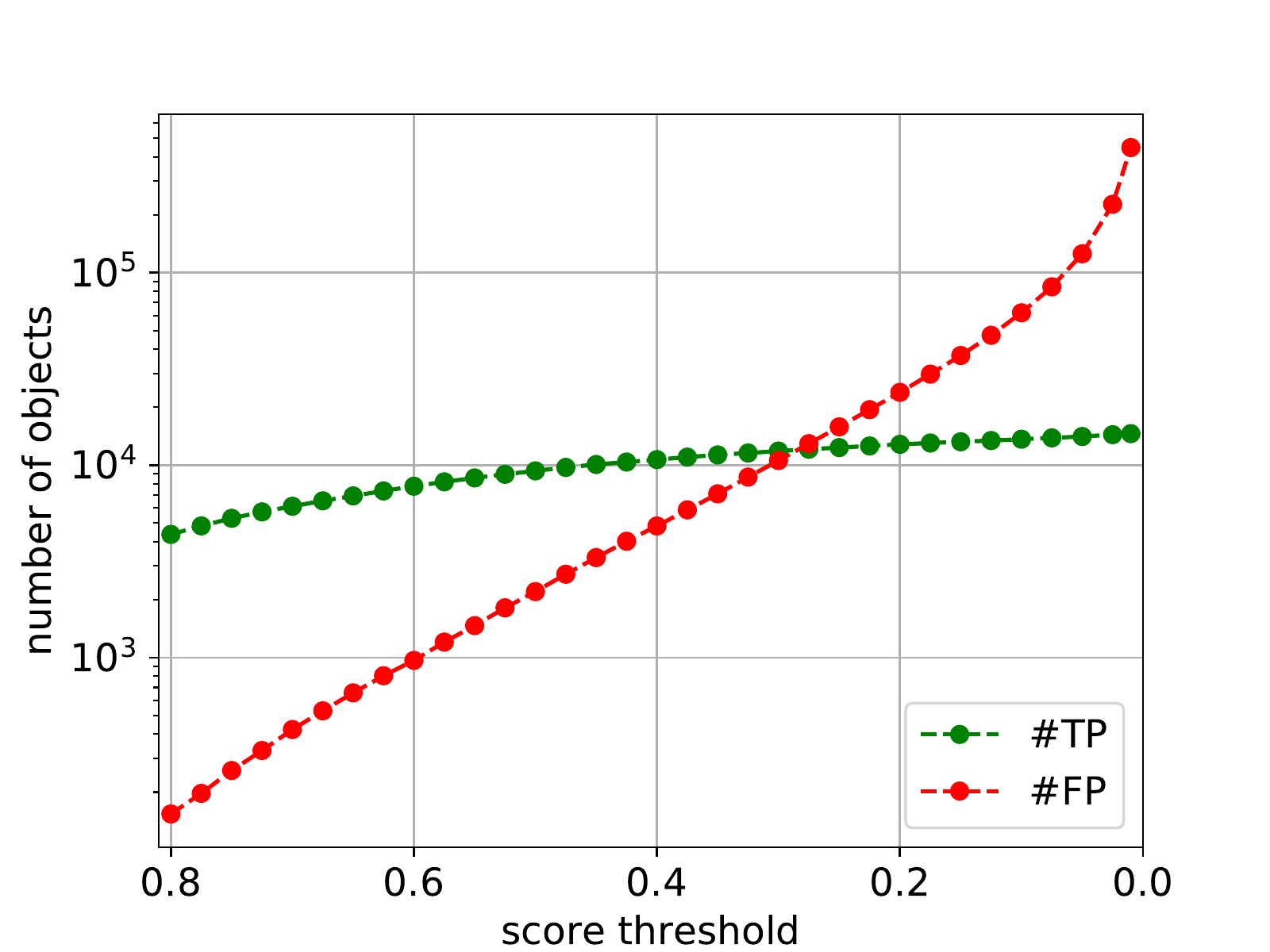}
  \caption{COCO + YOLOv3}
  \label{fig:3}
\end{subfigure}

\medskip
\begin{subfigure}{0.33\textwidth}
  \includegraphics[width=\textwidth]{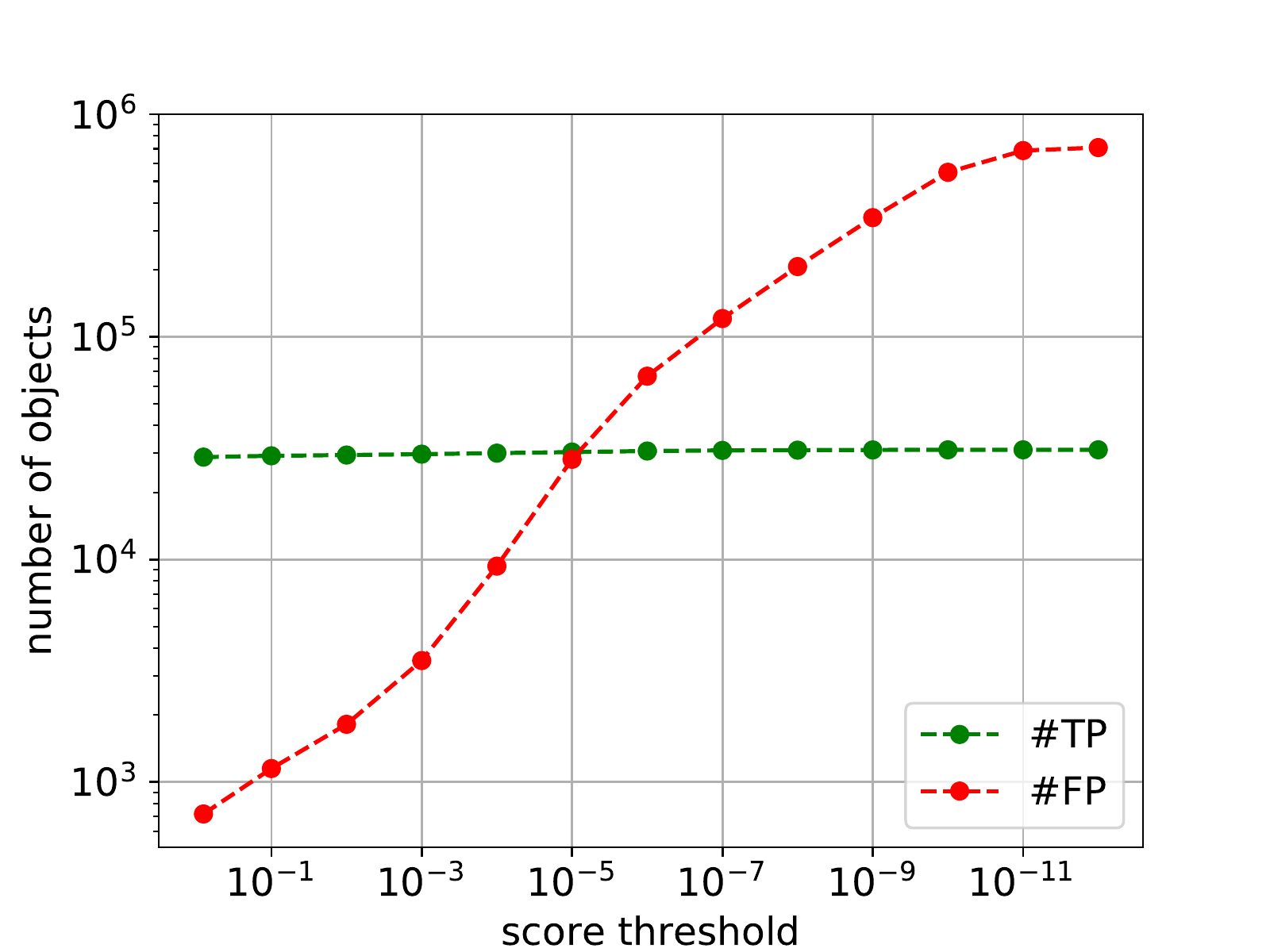}
  \caption{Kitti + Faster-RCNN}
  \label{fig:4}
\end{subfigure}\hfil 
\begin{subfigure}{0.33\textwidth}
  \includegraphics[width=\textwidth]{figs/tp_fp_real_voc.pdf}
  \caption{Pascal VOC + Faster-RCNN}
  \label{fig:5}
\end{subfigure}\hfil 
\begin{subfigure}{0.33\textwidth}
  \includegraphics[width=\textwidth]{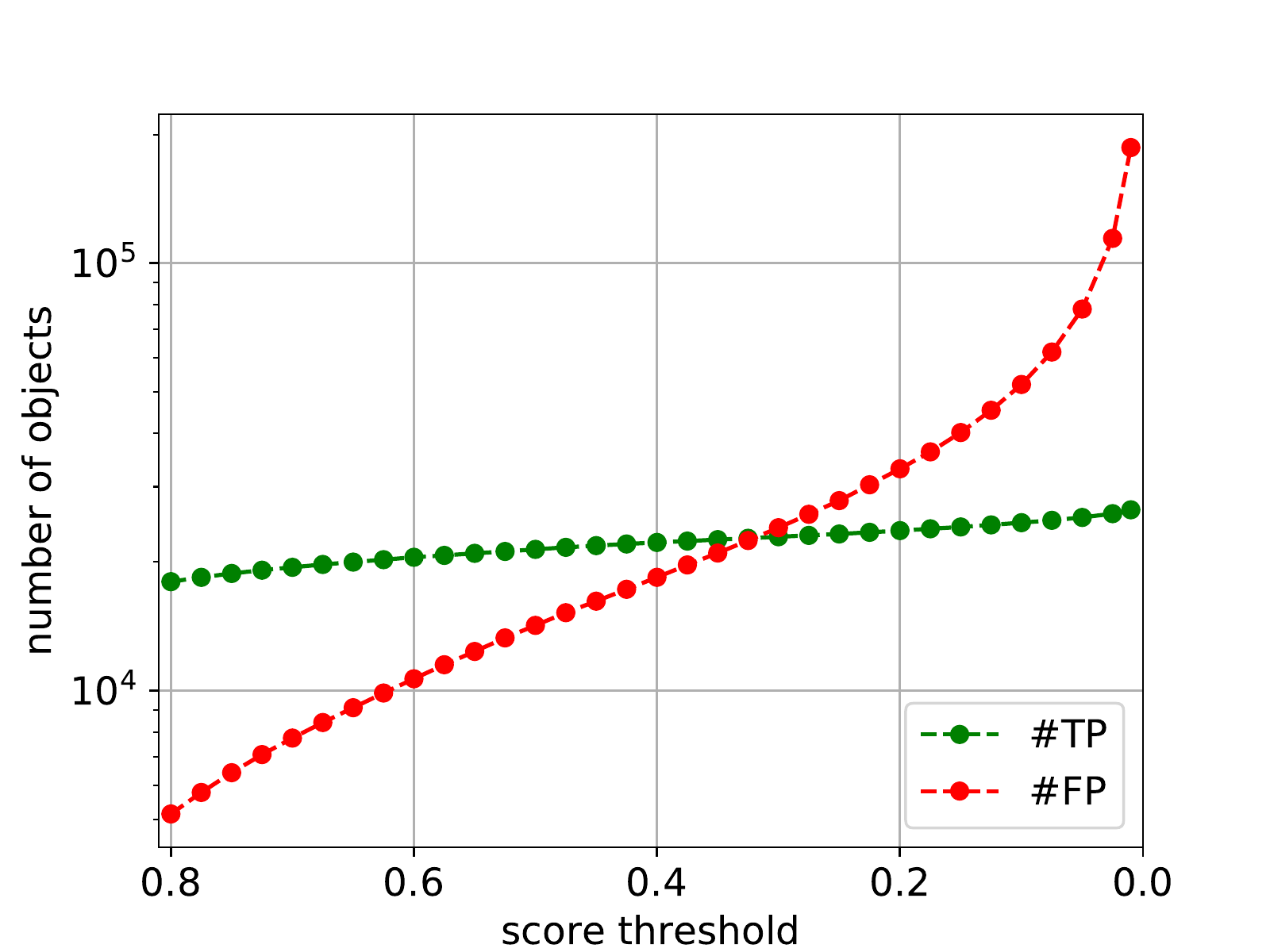}
  \caption{COCO + Faster-RCNN}
  \label{fig:6}
\end{subfigure}
\caption{Comparison of the number of true positives (TP) and false positives (FP) for different score thresholds, the YOLOv3 and Faster-RCNN network and the Kitti, Pascal VOC and COCO datasets.}
\label{fig:images}
\end{figure*}

\Cref{fig:images} depicts the number of true positives and false positives for the YOLOv3 and Faster-RCNN network and the Kitti, Pascal VOC and COCO datasets. It is intuitively clear that as the score threshold decreases the number TPs and FPs increases. The sum of TPs and FPs equals the number of all predicted bounding boxes and therefore constitutes the number of examples for training and evaluation of our meta models. The order of magnitude of the number of predictions is between $10^3$ and $10^6$.

In what follows, all results (if not stated otherwise)
were computed from $10$ repeated runs where training and validation sets (both of the same size) were re-sampled. We give mean results as well as standard deviations over the obtained results in brackets.

\begin{table}[t]
\scalebox{.8}{{\setlength{\extrarowheight}{5.5pt}%
\begin{tabular}{|l|r||l|r|}
\cline{1-4}
$s_{\mathit{std}}^{(i)}$                       &    0.8370   &  $s_{\mathit{mean,MC}}^{(i)}$                   &    0.8239 \\
\cline{1-4}
$s^{(i)}$                           &    0.8231   &  $s_{\mathit{mean}}^{(i)}$                      &    0.8147 \\
\cline{1-4}
$N^{(i)}$                               &    0.7149   &  $\IoU_{\mathit{pb},\mathit{std}}^{(i)}$                      &    0.6529 \\
\cline{1-4}
$\IoU_{\mathit{pb},\mathit{high}}^{(i)}$                      &    0.6021   &  $\IoU_{\mathit{pb},\mathit{mean}}^{(i)}$                     &    0.5019 \\
\cline{1-4}
$\hat{r}^{(i)}_{\mathrm{max,\mathit{high}}}$                           &    0.4174   &  $\hat{r}^{(i)}_\mathrm{max}$                               &    0.4054 \\
\cline{1-4}
$d/g_{\mathit{high}}^{(i)}$                 &    0.4049   &  $\hat{r}^{(i)}_{\mathrm{max,\mathit{mean,MC}}}$                       &    0.4044 \\
\cline{1-4}
$x_{\mathit{std}}^{(i)}$                           &    0.4029   &  $\hat{r}^{(i)}_{\mathrm{max,\mathit{{std}}}}$                           &    0.4013 \\
\cline{1-4}
$d/g_{\mathit{std}}^{(i)}$                 &    0.3995    &  $g_{\mathit{std}}^{(i)}$                      &    0.3904 \\
\cline{1-4}
$\hat{r}^{(i)}_{\mathrm{max,\mathit{{mean}}}}$                          &  0.3876     & $g_{\mathit{high}}^{(i)}$                        &  0.3821 \\
\cline{1-4}
\end{tabular}
}}
{\caption{Strongest Pearson correlation coefficients for some constructed box-wise metrics for the Kitti dataset, the YOLOv3 network and a score threshold $t=0.01$.   \label{tab:corr_coeff}}%
}
\end{table}

\paragraph{Correlation of box-wise metrics with the $\boldsymbol{\IoU}$.}
\Cref{tab:corr_coeff} contains the Pearson correlation coefficients of the box-wise metrics with the $\IoU$ of prediction and ground truth for the Kitti test images and a score threshold $t=0.01$ for the YOLOv3 network. The score metrics seem to have strong correlations with the $\IoU$, which is 
expected as it is supposed to discriminate true positives from false positives.
The endings $\mathit{high},\mathit{mean},\mathit{std}$ represent the maximum, the arithmetic mean and the standard deviation, respectively, of the corresponding filtered candidate boxes for a given metric. 
The ending $\mathit{mean,MC}$ represents the arithmetic mean of the corresponding dropout predictions $\hatbox, \hatbox_1,...,\hatbox_J$ for the respective metric.
Note that, although the $4$ score related metrics ($s_\mathit{std}^{(i)},\ s_\mathit{mean,MC}^{(i)},\ s^{(i)},\ s_\mathit{mean}^{(i)}$) show the highest correlation with the $\IoU \geq 0.8$, these metrics may be very similar. 
Indeed, the correlation coefficients between these four metrics range from 0.81 to 0.99.
Four additional metrics ($N^{(i)},\ \IoU_{pb_{std}}^{(i)},\ \IoU_{\mathit{pb},\mathit{high}}^{(i)},\ \IoU_{\mathit{pb},\mathit{mean}}^{(i)}$) also show decent correlations $0.5 \leq \rho \leq 0.8$, all other metrics only show a minor correlation. However they may still contribute to a diverse set of metrics.

\begin{table*}[t]
\centering
\scalebox{0.835}{{\setlength{\extrarowheight}{0.75pt}%
\begin{tabular}{||l|l|l|l||c|c|c||}
\cline{1-7}
\multicolumn{7}{||c||}{Meta Regression $\IoU$} \\
\cline{1-7}
Network & Dataset & Score threshold $t$ & Method & Score baseline & MetaDetect & MetaDetect+Dropout \\
\cline{1-7}
\multirow{9}{*}{YOLOv3} & \multirow{9}{*}{Kitti} & \multirow{3}{*}{$0.5$} & $LR$ & $0.4188(\pm0.0080)$  & $0.4447(\pm0.0059)$ & $0.4477(\pm0.0065)$ \\
 & &  & $GB$ & $0.3966(\pm0.0069)$ & $0.4485(\pm0.0104)$ & $0.4478(\pm0.0102)$   \\
  & &  & $NN$ & $0.3448(\pm0.0096)$ & $0.4435(\pm0.0097)$ & $0.4436(\pm0.0114)$   \\
\cline{3-7}
 &  & \multirow{3}{*}{$0.3$} & $LR$ & $0.5837(\pm0.0097)$ & $0.6131(\pm0.0094)$ & $0.6116(\pm0.0098)$ \\
 & &  & $GB$ & $0.5679(\pm0.0099)$ & $0.6206(\pm0.0095)$ & $0.6216(\pm0.0084)$ \\
  & &  & $NN$ & $0.5704(\pm0.0088)$ & $0.6078(\pm0.0094)$ & $0.6166(\pm0.0096)$ \\
\cline{3-7}
 &  & \multirow{3}{*}{$0.1$} & $LR$ & $0.7133(\pm0.0032)$ & $0.7594(\pm0.0032)$ & $0.7591(\pm0.0033)$   \\
 & &  & $GB$ & $0.7138(\pm0.0037)$ & $0.7766(\pm0.0025)$ & $0.7780(\pm0.0025)$  \\
  & &  & $NN$ & $0.7120(\pm0.0029)$ & $0.7568(\pm0.0042)$ & $0.7635(\pm0.0060)$  \\
\cline{1-7}
\end{tabular}   
}}
\caption{Comparison of $R^2$ values for the score baseline and all available metrics (with and without dropout) for Kitti and YOLOv3. We used linear regression (LR), gradient boosting (GB) and shallow neural nets (NN) for the task of meta regression.
}
\label{tab:meta_regression}
\end{table*}

\begin{figure*}[t]
\begin{floatrow}
\ffigbox[0.475\textwidth]{%
  \includegraphics[width=.47\textwidth]{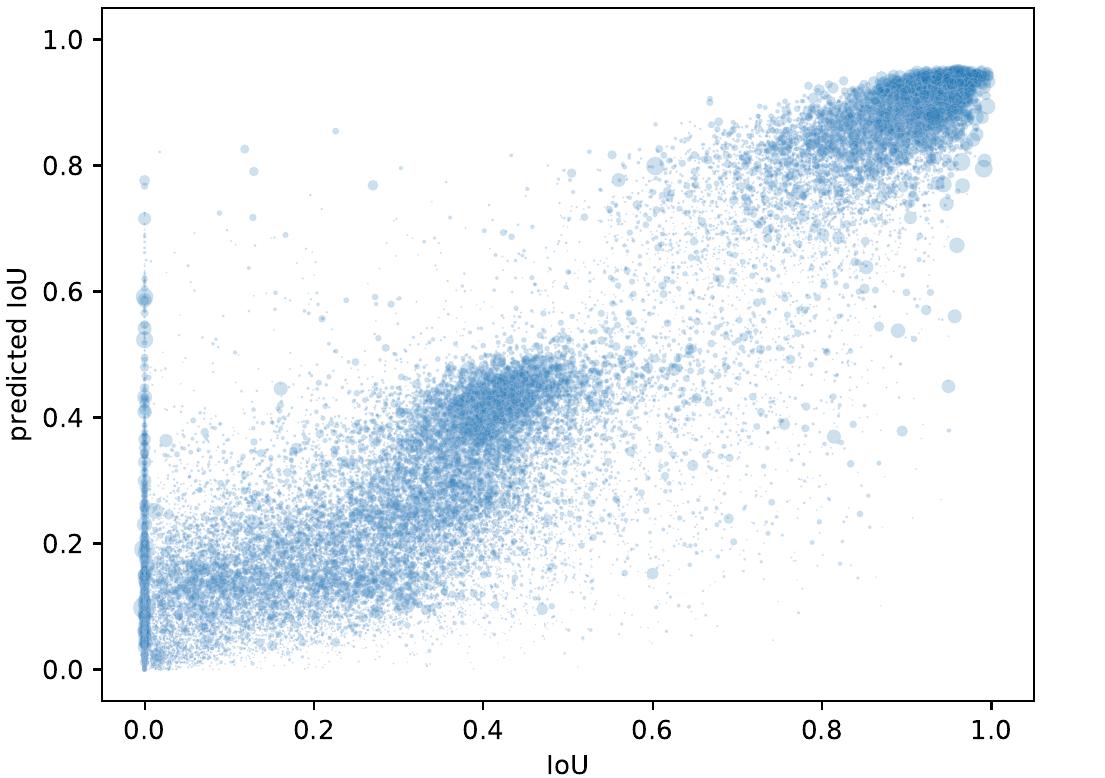}
}{%
  \caption{Box-wise scatter plot of true $\IoU$ and predicted $\IoU$ values for the Kitti dataset, the YOLOv3 network and a score threshold $t=0.01$. The predicted $\IoU$ values are generated with gradient boosting. \label{fig:illus3}
  }
}
\ffigbox[0.475\textwidth]{%
  \includegraphics[width=.5\textwidth]{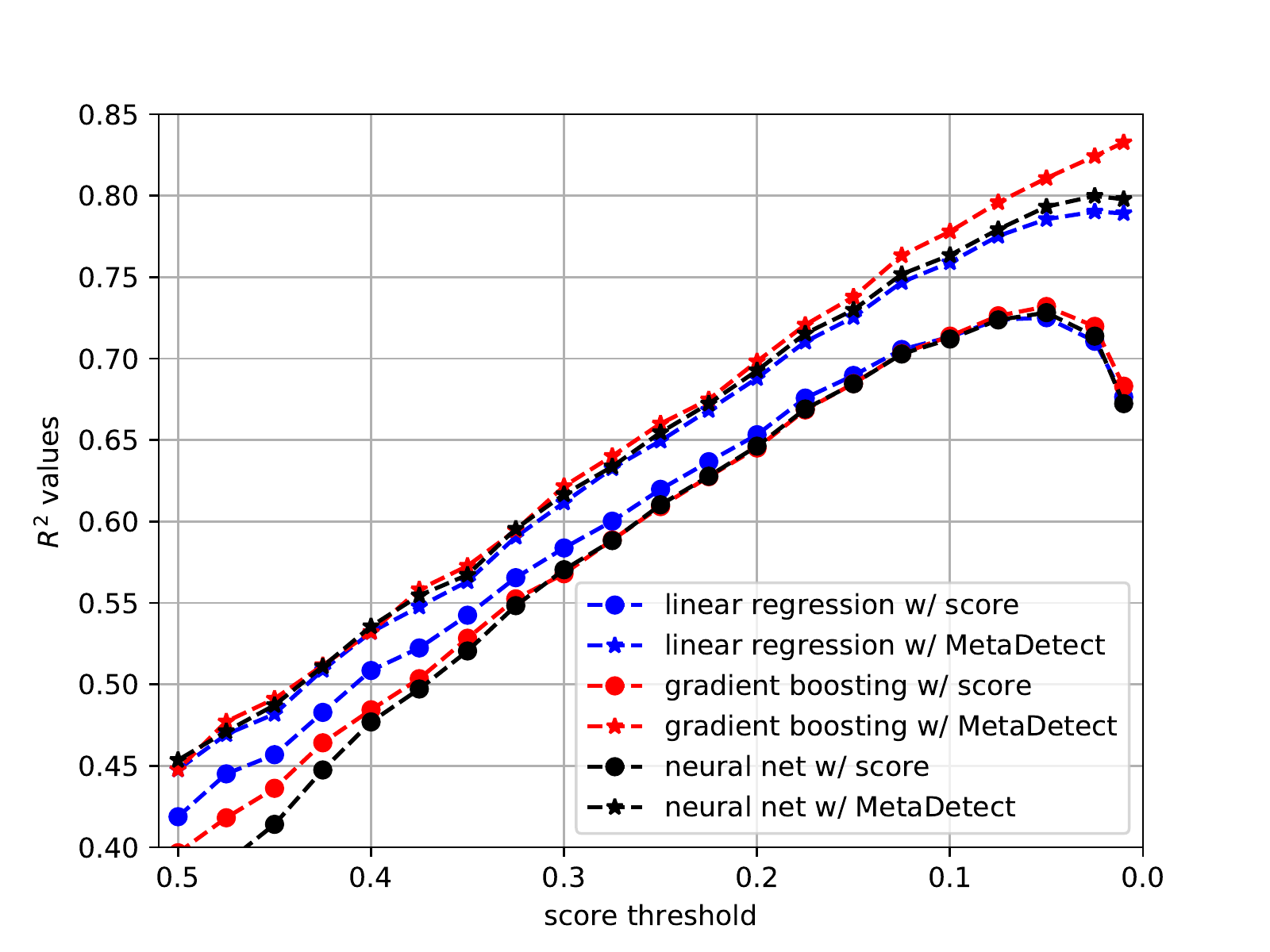}
}{%
  \caption{$R^2$ values for the task of meta regression for the score baseline and all uncertainty measures for the Kitti dataset, the YOLOv3 network and different score values. The $R^2$ values are calculated with linear regression, gradient boosting and a shallow neural net. \label{fig:illus4}
  }
}
\end{floatrow}
\end{figure*}

\paragraph{Comparison of different meta classifiers and regressors.}

For meta classification (classification of $\IoU\geq0.5$ vs.\ $\IoU<0.5$), we compare results in terms of classification accuracy and in terms of the area under curve corresponding to the receiver operator characteristic curve (AUROC, see~\cite{DBLP:conf/icml/DavisG06}). The receiver operator characteristic curve is obtained by varying the decision threshold of the classification output for deciding whether $\IoU\geq0.5$ or $\IoU<0.5$. For the task of meta regression we state resulting standard deviations $\sigma$ of the linear regression fit's residual as well as $R^2$ values.
Throughout this section, we consider the following combinations of inputs:
Score / score baseline refers to a meta regressor or classifier that was only trained by means of the score metric $s^{(i)}$ which is a sensible baseline. MetaDetect refers to a meta regressor or classifier trained with $46+C$ metrics that include all metrics except for those that correspond to MC dropout. MetaDetect+Dropout refers to all available $66+C$ metrics including dropout. MetaDetect+Dropout is only available for YOLOv3, since for this network we inserted a dropout layer with dropout rate $0.5$ in all three detection branches right before the final convolutional layer and retrained the network.

For the task of meta regression and meta classification several models (meta models) can be used. The meta models we consider 
are linear regression (LR), gradient boosting (GB) and a shallow neural network (NN) with two hidden layers. In~\cref{tab:meta_regression} we state $R^2$-values for meta regression with the Kitti dataset and the YOLOv3 network for three different score thresholds $t$. We be observed that gradient boosting outperforms linear regression and the shallow neural network consistently for all three score thresholds. This is in accordance to the results depicted by~\cref{fig:illus4}. Gradient boosting outperforms the linear regression and the shallow neural network as a meta regression model for all 33 different score thresholds. In all our tests, we observed the same behavior for meta regression and meta classification for all three datasets. For this reason, only meta regression and meta classification results with gradient boosting are presented in the following. These findings indicate that there is a significant portion of mutual information contained in our metrics. The increase in $R^2$ when going from a single score metric to all our $46+C$ metrics is ranging from $1.52$ to $9.83$ percent points (pp). On average the gain is about 4.86 pp.
When further extending the considered metrics to $66+C$ by the MC dropout metrics, we do neither observe a significant nor consistent increase in $R^2$ values.

\paragraph{Comparisons for different datasets, networks and sets of metrics.}

Due to our observations in the previous paragraph, we fix all our meta classifies and regressors to gradient boosting.
\Cref{tab:meta_regression_gb} presents results for meta regression in terms of regression $R^2$ for all three datasets and both networks with three score thresholds each. Obviously, the tendencies indicated by the previously studied \cref{tab:meta_regression} are confirmed by \cref{tab:meta_regression_gb} for the different datasets and networks. In all cases we MetaDetect provides distinct increases in comparison to the score baseline. For the COCO dataset, YOLOv3 network and score threshold $0.1$ we even observe an increase of $9.83$ percent points. On average, the incease amounts to $4.86$ percent points. Analogously to our findings in the previous paragraph, MetaDetect+Dropout is not able to further improve. In our tests we also observed, although the dropout rate is set to $0.5$, that the variation introduced by dropout inference seems to be rather limited. We performed $10$ forward passes under dropout. In general, a more stochastic behavior of the inference, which could be promoted by additional dropout layers of an analogous batch normalization approach, could lead to an improvement. Also here, we refer to the modular nature of our framework MetaDetect which allows for the incorporation of any uncertainty metric.

\begin{table*}[t]
\centering
\scalebox{0.835}{{\setlength{\extrarowheight}{0.75pt}%
\begin{tabular}{||l|l|l||c|c|c||}
\cline{1-6}
\multicolumn{6}{||c||}{Meta Regression $\IoU$} \\
\cline{1-6}
Network & Dataset & Score threshold $t$ & Score baseline & MetaDetect & MetaDetect+Dropout \\
\cline{1-6}
\multirow{9}{*}{YOLOv3} & \multirow{3}{*}{Kitti} & $0.5$ & $0.3966(\pm0.0069)$ & $0.4485(\pm0.0104)$ & $0.4478(\pm0.0102)$   \\
\cline{3-6}
 & & $0.3$ & $0.5679(\pm0.0099)$ & $0.6206(\pm0.0095)$ & $0.6216(\pm0.0084)$ \\
\cline{3-6}

 & & $0.1$ & $0.7138(\pm0.0037)$ & $0.7766(\pm0.0025)$ & $0.7780(\pm0.0025)$  \\
\cline{2-6}
 & \multirow{3}{*}{Pascal VOC} & $0.5$ & $0.3468(\pm0.0094)$ & $0.3853(\pm0.0113)$ & $0.3851(\pm0.0098)$   \\
\cline{3-6}
 & & $0.3$ & $0.4234(\pm0.0081)$ & $0.4700(\pm0.0078)$ & $0.4672(\pm0.0089)$  \\
\cline{3-6}
 & & $0.1$& $0.5384(\pm0.0051)$ & $0.5907(\pm0.0057)$ & $0.5879(\pm0.0065)$   \\
\cline{2-6}
 & \multirow{3}{*}{COCO} & $0.5$ & $0.2317(\pm0.0089)$ & $0.2787(\pm0.0077)$ & $0.2774(\pm0.0084)$   \\
\cline{3-6}
 & & $0.3$ & $0.4428(\pm0.0058)$ & $0.4978(\pm0.0070)$ & $0.4960(\pm0.0062)$ \\
\cline{3-6}
 & & $0.1$ & $0.4931(\pm0.0028)$ & $0.5914(\pm0.0030)$ & $0.5911(\pm0.0028)$   \\
\cline{1-6}
\multirow{9}{*}{Faster-RCNN} & \multirow{3}{*}{Kitti} & $10^{-1}$ & $0.3703(\pm0.0136)$ & $0.4101(\pm0.0153)$ &   \\
\cline{3-6}
 & & $10^{-6}$ & $0.8930(\pm0.0017)$ & $0.9178(\pm0.0014)$ &    \\
\cline{3-6}
 & & $10^{-12}$ & $0.7927(\pm0.0020)$ & $0.8819(\pm0.0013)$ &  \\
\cline{2-6}
 & \multirow{3}{*}{Pascal VOC} & $0.5$ & $0.4844(\pm0.0106)$ & $0.5430(\pm0.0121)$ &  \\
\cline{3-6}
 & & $0.3$ & $0.5682(\pm0.0084)$ & $0.6182(\pm0.0089)$ &   \\
\cline{3-6}
 & & $0.1$ & $0.6289(\pm0.0054)$ & $0.6836(\pm0.0074)$ &  \\
\cline{2-6}
 & \multirow{3}{*}{COCO} & $0.5$ &  $0.4178(\pm0.0046)$ & $0.4330(\pm0.0052)$ &   \\
\cline{3-6}
 & & $0.3$ & $0.4730(\pm0.0076)$ & $0.4890(\pm0.0068)$ &  \\
\cline{3-6}
 & & $0.1$ &  $0.5438(\pm0.0056)$ & $0.5651(\pm0.0041)$ &   \\
\cline{1-6}
\end{tabular}   
}}
\caption{Comparison of $R^2$ values for the score baseline and all available metrics (with and without dropout) for Kitti, Pascal VOC and COCO. Especially, for Faster-RCNN there are no dropout metrics available. We used gradient boosting (GB) for the task of meta regression.
}
\label{tab:meta_regression_gb}
\end{table*}

\Cref{fig:illus3} shows a scatter plot of the true $\IoU$ of prediction and ground truth and the $\IoU$ estimated by MetaDetect. The scattered points a well concentrated along the diagonal axis corresponding to the identity $\mathrm{id}: a \mapsto a $. This signals, that we obtain a well calibrated $\IoU$ estimate. The limited deviation from the identity shows, that is estimate gives rise to predictive uncertainty and an object-wise quality estimate. For example images, we refer to \cref{fig:Collage_Regression}


Considering the task of meta classification which amounts to false positive detection, we also achieve clear improvements when considering all our metrics instead of only considering the score $s^{(i)}$.
In this task we can also consider another baseline which is given just by random guessing. In case of a balanced dataset with equal amount of false positives and false negatives, both the classification accuracy and the AUROC have values of $0.5$ in the case of random guessing. 

Results are summarized in \cref{tab:meta_classification} in terms of classification accuracy and in \cref{tab:meta_classification_AUROC} in terms of classification AUROC. Since different score thresholds $t$ amount to different ratios of false positives and true positives, we apply SMOTE \cite{SMOTE} in order to balance the meta classification dataset. Our findings in these tables are similar to those for meta regression, the advantage of MetaDetect over the score baseline seems to be slightly more pronounced in terms of AUROC than in terms of accuracy. In terms of accuracy, we observe an average increase of 2.07 pp, while in terms of AUROC, we obtain an average increase of 2.32 pp. The highest increase overall is obtained for the YOLOv3 network and the COCO dataset with a score threshold $t=0.5$ with $7.09$ pp in terms of classification accuracy and for the YOLOv3 network and the COCO dataset with a score threshold $t=0.5$ with $9.18$ pp in terms of AUROC.

\begin{figure*}[t]
\begin{minipage}[c]{.99\textwidth}
\begin{center}
\includegraphics[width=.13\textwidth]{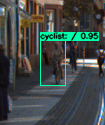}
\includegraphics[width=.13\textwidth]{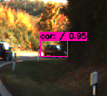}
\includegraphics[width=.13\textwidth]{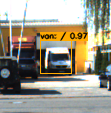}
\includegraphics[width=.13\textwidth]{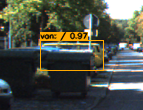}
\includegraphics[width=.13\textwidth]{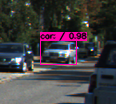}
\includegraphics[width=.13\textwidth]{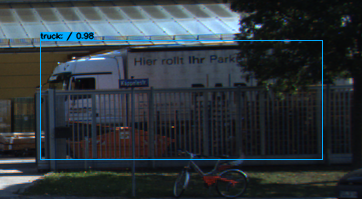}
\includegraphics[width=.13\textwidth]{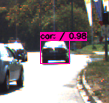} \\
\includegraphics[width=.13\textwidth]{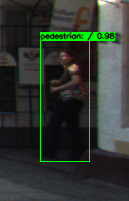}
\includegraphics[width=.13\textwidth]{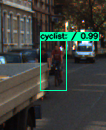}
\includegraphics[width=.13\textwidth]{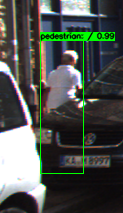}
\includegraphics[width=.13\textwidth]{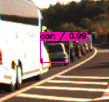}
\includegraphics[width=.13\textwidth]{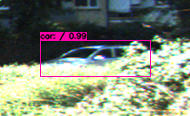}
\includegraphics[width=.13\textwidth]{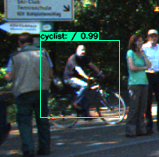}
\includegraphics[width=.13\textwidth]{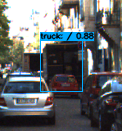}
\end{center}
\captionof{figure}{Examples of predicted bounding boxes with a true $\IoU=0$ but with high meta classification probabilities. These examples predicted by the YOLOv3 network are counted as FPs as there is no corresponding ground truth available. Our prediction quality estimates signal with very high $\IoU$ values, that an object is present.
The predictions, and therefore the data to train and evaluate gradient boosting, are 
obtained by the YOLOv3 network 
applied to the Kitti dataset. \label{fig:Collage_Classification}
}
\end{minipage}
\end{figure*}

\Cref{fig:Collage_Classification} shows examples of predicted bounding boxes that have a true $\IoU=0$. Thus, according to the ground truth they are FPs, but with high meta classification probabilities. Indeed, these boxes each represent an object of the correct class. Hence, more reliable meta classifiers may also help to identify labeling errors.

\begin{table*}[t]
\centering
\scalebox{0.835}{{\setlength{\extrarowheight}{0.75pt}%
\begin{tabular}{||l|l|l||c|c|c||}
\cline{1-6}
\multicolumn{6}{||c||}{Meta Classification $\IoU\geq0.5$ vs. $\IoU<0.5$, Accuracies} \\
\cline{1-6}
Network & Dataset & Score threshold $t$  & Score baseline & MetaDetect & MetaDetect+Dropout \\
\cline{1-6}
\multirow{9}{*}{YOLOv3} & \multirow{3}{*}{Kitti} & $0.5$        & $0.8527(\pm0.0132)$ & $0.9042(\pm0.0145)$ & $0.9032(\pm0.0129)$   \\
\cline{3-6}
 & &  $0.3$                                                                            & $0.8905(\pm0.0075)$ & $0.9092(\pm0.0100)$ & $0.9090(\pm0.0091)$  \\
\cline{3-6}
 & &    $0.1$                                                                          & $0.9167(\pm0.0050)$ & $0.9207(\pm0.0058)$ & $0.9205(\pm0.0054)$      \\
\cline{2-6}
 & \multirow{3}{*}{Pascal VOC} & $0.5$                                                                          & $0.8106(\pm0.0097)$ & $0.8527(\pm0.0077)$ & $0.8537(\pm0.0072)$   \\
\cline{3-6}
 & & $0.3$                                                                             & $0.8259(\pm0.0077)$ & $0.8532(\pm0.0104)$ & $0.8524(\pm0.0103)$   \\
\cline{3-6}
 & &       $0.1$                                                                       & $0.8480(\pm0.0040)$ & $0.8628(\pm0.0048)$ & $0.8624(\pm0.0047)$   \\
\cline{2-6}
 & \multirow{3}{*}{COCO} & $0.5$                                                     & $0.7381(\pm0.0118)$ & $0.8090(\pm0.0112)$ & $0.8083(\pm0.0115)$    \\
\cline{3-6}
 & &  $0.3$                                                                            & $0.7882(\pm0.0066)$ & $0.8022(\pm0.0070)$ & $0.8023(\pm0.0069)$ \\
\cline{3-6}
 & &  $0.1$                                                                            & $0.8400(\pm0.0032)$ & $0.8861(\pm0.0036)$ & $0.8854(\pm0.0036)$   \\
\cline{1-6}
\multirow{9}{*}{Faster-RCNN} & \multirow{3}{*}{Kitti} & $10^{-1}$ & $0.9092(\pm0.0062)$ & $0.9266(\pm0.0060)$ &   \\
\cline{3-6}
 & &     $10^{-6}$                                                                        & $0.9678(\pm0.0017)$ & $0.9710(\pm0.0013)$ &     \\
\cline{3-6}
 & &    $10^{-12}$                                                                         & $0.9825(\pm0.0023)$ & $0.9892(\pm0.0004)$ &   \\
\cline{2-6}                                                                                     
 & \multirow{3}{*}{Pascal VOC} & $0.5$                                                      & $0.8270(\pm0.0043)$ & $0.8384(\pm0.0049)$ &     \\
\cline{3-6}
 & &    $0.3$                                                                        & $0.8420(\pm0.0042)$ & $0.8508(\pm0.0056)$ &    \\
\cline{3-6}
 & &   $0.1$                                                                          & $0.8642(\pm0.0040)$ & $0.8726(\pm0.0043)$ &    \\
\cline{2-6}                                                                                     
 & \multirow{3}{*}{COCO} & $0.5$                                                   & $0.7814(\pm0.0035)$ & $0.7865(\pm0.0036)$ &    \\
\cline{3-6}
 & &      $0.3$                                                                     & $0.7877(\pm0.0041)$ & $0.7913(\pm0.0029)$ &    \\
\cline{3-6}
 & &     $0.1$                                                                       & $0.8246(\pm0.0027)$ & $0.8430(\pm0.0032)$ &     \\
\cline{1-6}
\end{tabular}   
}}
\caption{Comparison of meta classification accuracies for the score baseline and all available metrics (with and without dropout) for the Kitti, Pascal VOC and COCO datasets. For Faster-RCNN there are no dropout metrics available. We used gradient boosting (GB) for meta classification.
}
\label{tab:meta_classification}
\end{table*}

\begin{table*}[t]
\centering
\scalebox{0.835}{{\setlength{\extrarowheight}{0.75pt}%
\begin{tabular}{||l|l|l||c|c|c||}
\cline{1-6}
\multicolumn{6}{||c||}{Meta Classification $\IoU\geq0.5$ vs. $\IoU<0.5$, AUROCs} \\
\cline{1-6}
Network & Dataset & Score threshold $t$ & Score baseline & MetaDetect & MetaDetect+Dropout \\
\cline{1-6}
\multirow{9}{*}{YOLOv3} & \multirow{3}{*}{Kitti} & $0.5$                                                                              & $0.9183(\pm0.0090)$ & $0.9635(\pm0.0065)$ & $0.9626(\pm0.0072)$   \\
\cline{3-6}
 & &       $0.3$                                                                       &  $0.9439(\pm0.0058)$ & $0.9655(\pm0.0049)$ & $0.9651(\pm0.0049)$ \\
\cline{3-6}
 & &       $0.1$                                                                       &  $0.9598(\pm0.0031)$ & $0.9717(\pm0.0025)$ & $0.9714(\pm0.0023)$      \\
\cline{2-6}                                                                    
 & \multirow{3}{*}{Pascal VOC} & $0.5$                                                                              &  $0.8772(\pm0.0082)$ & $0.9306(\pm0.0049)$ & $0.9304(\pm0.0040)$   \\
\cline{3-6}
 & &       $0.3$                                                                       &  $0.8950(\pm0.0072)$ & $0.9266(\pm0.0061)$ & $0.9265(\pm0.0057)$   \\
\cline{3-6}
 & &       $0.1$                                                                       &  $0.9183(\pm0.0035)$ & $0.9327(\pm0.0030)$ & $0.9326(\pm0.0027)$ \\
\cline{2-6}                                                                    
 & \multirow{3}{*}{COCO}  & $0.5$                                                                              &  $0.8063(\pm0.0145)$ & $0.8981(\pm0.0106)$ & $0.8979(\pm0.0098)$    \\
\cline{3-6}
 & &       $0.3$                                                                      &  $0.8578(\pm0.0070)$ & $0.8793(\pm0.0062)$ & $0.8793(\pm0.0062)$ \\
\cline{3-6}
 & &       $0.1$                                                                       &  $0.9125(\pm0.0027)$ & $0.9514(\pm0.0025)$ & $0.9515(\pm0.0020)$   \\
\cline{1-6}
\multirow{9}{*}{Faster-RCNN} & \multirow{3}{*}{Kitti} & $10^{-1}$                                                                            &  $0.9580(\pm0.0051)$ & $0.9694(\pm0.0036)$ &  \\
\cline{3-6}
 & &       $10^{-6}$                                                                      &  $0.9891(\pm0.0009)$ & $0.9949(\pm0.0004)$ &    \\
\cline{3-6}
 & &       $10^{-12}$                                                                      &  $0.9982(\pm0.0002)$ & $0.9993(\pm0.0001)$ &    \\
\cline{2-6}                                                                                     
 & \multirow{3}{*}{Pascal VOC} & $0.5$                                                                            &  $0.8930(\pm0.0036)$ & $0.9090(\pm0.0036)$ &     \\
\cline{3-6}
 & &       $0.3$                                                                      &  $0.9113(\pm0.0036)$ & $0.9196(\pm0.0034)$ &   \\
\cline{3-6}
 & &       $0.1$                                                                      &  $0.9309(\pm0.0028)$ & $0.9399(\pm0.0028)$ &    \\
\cline{2-6}                                                                                     
 & \multirow{3}{*}{COCO} & $0.5$                                                                             &  $0.8482(\pm0.0022)$ & $0.8608(\pm0.0032)$ &    \\
\cline{3-6}
 & &       $0.3$                                                                      &  $0.8623(\pm0.0033)$ & $0.8669(\pm0.0028)$ &    \\
\cline{3-6}
 & &       $0.1$                                                                     &  $0.8979(\pm0.0021)$ & $0.9157(\pm0.0022)$ &     \\
\cline{1-6}
\end{tabular}   
}}
\caption{Comparison of Meta-Classification-AUROC values for the score baseline and all available metrics (with and without dropout) for the Kitti, Pascal VOC and COCO datasets. For Faster-RCNN there are no dropout metrics available since the pre-trained network does not use dropout. We used gradient boosting (GB) for meta classification.
}
\label{tab:meta_classification_AUROC}
\end{table*}

\section{Conclusion and Outlook}

In this work we extended the notion of~\cite{A_Baseline} from classification to object detection, considering the tasks of meta classification and meta regression introduced in \cite{MetaSeg,NestedMetaSeg,Time_Dynamic_MetaSeg} for semantic segmentation. We introduced a generic framework for quality estimation and false positive detection that can be extended by any uncertainty measure for object detection. To demonstrate the ability of our framework to perform these tasks more efficiently than state-of-the-art baselines, we introduced a variety of different uncertainty metrics and also considered widely used MC dropout \cite{MC-Dropout} uncertainty. To study the individual metric's performance we compared correlation coefficients of the different metrics with each other and found that some of our metrics may well contribute to the overall meta regression and classification performance. This is confirmed by further numerical experiments where we compared different meta regressors and classifiers, i.e., logistic/linear regression, gradient boosting and shallow neural networks. We found that gradient boosting yields the best performance. Furthermore, with a thorough study of meta classification and regression performance over three different datasets and two different object detection networks, our method consistently outperforms the score baseline by a significant margin. In terms of meta classification we improve over the results of the score baseline by up to $7.09$ pp classification accuracy and $9.18$ pp AUROC.
For meta regression obtain an improvement up to $9.83$ pp in $R^2$.

We plan to incorporate these improved quality estimates into an active learning pipeline as well as performing an evaluation of the applicability of MetaDetect to data quality estimation. We believe that many labeling erros can be detected by means of a well-calibrated quality estimate.

We make our source code publicly available at \url{https://github.com/schubertm/MetaDetect}.



\paragraph{Acknowledgements.}
We thank Hanno Gottschalk for discussion and useful advice. Furthermore, we acknowledge support by the European Regional Development Fund (ERDF), grant-no.\ EFRE-0400216. \\
\includegraphics[width=.5\linewidth]{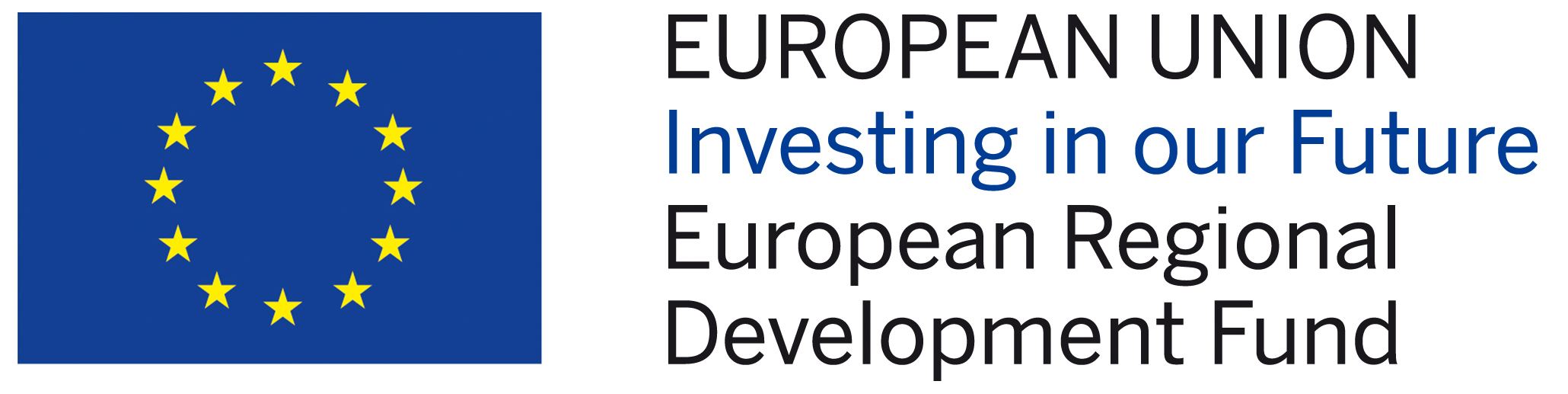}

{\small
\bibliographystyle{ieee}
\bibliography{biblio}
}

\end{document}